%% file: eccv2022submission.tex
\documentclass[runningheads]{llncs}
\usepackage{graphicx}

\usepackage{tikz}
\usepackage{comment}
\usepackage{booktabs}
\usepackage{amsmath,amssymb} 
\usepackage{color}
\usepackage[colorlinks,linkcolor=red]{hyperref}
\usepackage[misc]{ifsym}
\usepackage[capitalize]{cleveref}
\usepackage{algpseudocode}
\usepackage{multirow}
\usepackage{multicol}
\usepackage{makecell}
\usepackage{enumitem}
\usepackage{float}
\usepackage{subfigure}
\usepackage{wrapfig}
\newcommand{\etal}[0]{\emph{et al. }}
\usepackage[accsupp]{axessibility}  

\begin{document}
\pagestyle{headings}
\mainmatter
\def\ECCVSubNumber{4243}  

\title{Unifying Event Detection and Captioning as Sequence Generation via Pre-Training}

\titlerunning{Unifying Event Detection and Captioning as Sequence Generation}

\author{Qi Zhang \and
Yuqing Song \and
Qin Jin\thanks{Corresponding author}
}

\authorrunning{Q. Zhang et al.}
\institute{School of Information, Renmin University of China \\
\{zhangqi1996, syuqing, qjin\}@ruc.edu.cn
}

\maketitle

\begin{abstract}
Dense video captioning aims to generate corresponding text descriptions for a series of events in the untrimmed video, which can be divided into two sub-tasks, event detection and event captioning. Unlike previous works that tackle the two sub-tasks separately, recent works have focused on enhancing the inter-task association between the two sub-tasks. However, designing inter-task interactions for event detection and captioning is not trivial due to the large differences in their task specific solutions. Besides, previous event detection methods normally ignore temporal dependencies between events, leading to event redundancy or inconsistency problems. To tackle above the two defects, in this paper, we define event detection as a sequence generation task and propose a unified pre-training and fine-tuning framework to naturally enhance the inter-task association between event detection and captioning. Since the model predicts each event with previous events as context, the inter-dependency between events is fully exploited and thus our model can detect more diverse and consistent events in the video. Experiments on the ActivityNet dataset show that our model outperforms the state-of-the-art methods, and can be further boosted when pre-trained on extra large-scale video-text data. Code is available at \url{https://github.com/QiQAng/UEDVC}.
\end{abstract}

\input{subsection/introduce}
\input{subsection/related_work}
\input{subsection/method}
\input{subsection/experiment}

\input{subsection/conclusion}

\noindent{}\textbf{Acknowledgement.} 
This work was partially supported by National Key R\&D Program of China (No. 2020AAA0108600) and National Natural Science Foundation of China (No. 62072462).

\clearpage
%
%
\bibliographystyle{splncs04}
\bibliography{egbib}
\end{document}

%% file: subsection/introduce.tex
\section{Introduction}
Dense Video Captioning (DVC)~\cite{Krishna_2017_ICCV}, as one of the important tasks in video understanding, aims to localize and describe multiple events in untrimmed videos.
The early mainstream approaches \cite{2020Streamlined,audio_dvc_single,Iashin_2020_CVPR_Workshops_single,Yang_2018_dvc_single,wang2021echr} normally decompose the DVC task into two sub-tasks, event detection and event captioning, and tackle the two sub-tasks separately.
However, an obvious limitation of these methods is that they ignore the association of the two sub-tasks which can benefit from each other.

To address this limitation, some recent works \cite{Wang_2018_CVPR_tda_cg,Deng_2021_CVPR,Zhou_2018_CVPR_MT,Li_2018_CVPR_dvc,Wang_2021_ICCV_pdvc} explore to enhance the inter-task association between event detection and event captioning. For example, 
Deng \etal \cite{Deng_2021_CVPR} propose a "top-down" framework which connects visual and language information to localize events or connects visual and event information to generate captions at different stages to enforce the interaction between the two sub-tasks.
Wang \etal \cite{Wang_2021_ICCV_pdvc} propose to share the same intermediate features and jointly optimize the two sub-tasks.
Although the mutual promotion between the two sub-tasks has been witnessed, it is not trivial to design inter-task interactions for event detection and event captioning due to the large differences in their task specific solutions, which makes them hard to fully benefit from each other.

\begin{figure}[t]
	\begin{center}
		\includegraphics[width=\columnwidth]{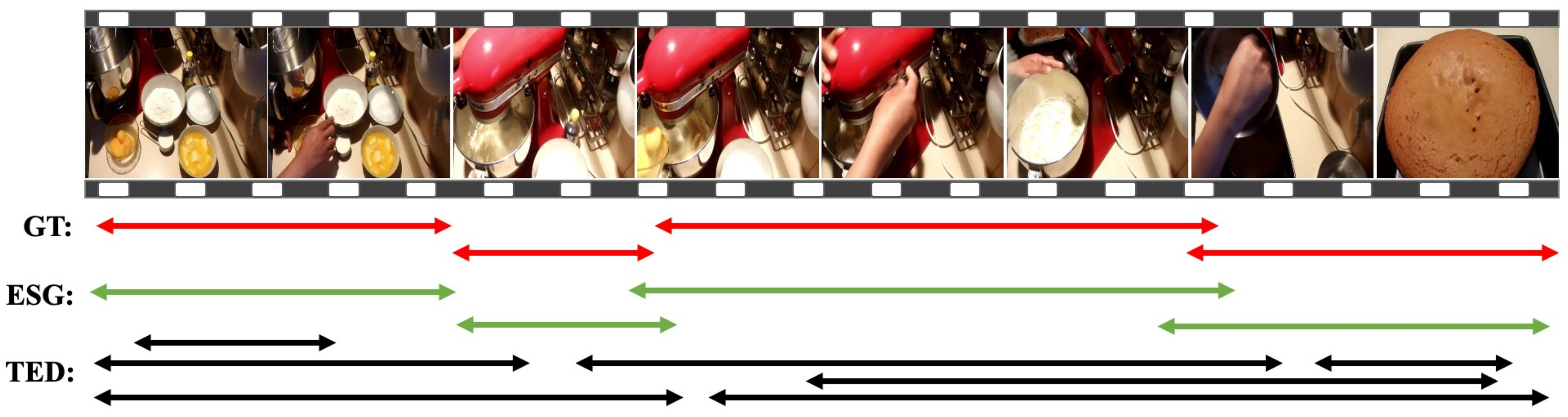}
	\end{center}

	\caption{Illustration of event detection results by the traditional event detection method (TED for short) and our proposed event sequence generation method (ESG for short). The TED usually predicts redundant events with a high degree of overlaps due to the lack of event relationship modeling, while our ESG can generate events sequentially with previous events as context. GT refers to the ground-truth event annotation. }

	\label{fig:1}
\end{figure}

Besides, the previous methods detect the event boundaries solely based on the video information or the hybrid knowledge of video and text, which ignores the temporal relationships between multiple events.
It easily leads to redundant event detection, either producing a large number of event candidates, or having excessive overlap between events.
As shown in Figure~\ref{fig:1}, due to the lack of consideration of the temporal relationship between events, the traditional event detection methods usually detect redundant events with a high degree of overlap.
In fact, there are strong temporal dependencies between events in the video. For example, in the instructional videos, events usually occur one after another, explaining each operation step.

To address the above mentioned two limitations, in this paper, we propose to define event detection as a sequence generation task, which generates the event sequence one by one conditioned on previously detected events.
It can fully exploit the previous events as context to avoid generating redundant events, and unify the event detection and event captioning sub-tasks into the same framework to explore inter-task interactions in a simpler but highly effective way.
Specifically, we define the event as a new modality and propose a unified video-event-text pre-training and fine-tuning framework based on the transformer architecture~\cite{vaswani:transformer} for both the event detection and the event captioning.
We employ two pre-training tasks including Masked Language Modeling (MLM) and Masked Video Feature Regression (MVFR) to learn the video-text representation for event captioning, and propose a novel pre-training task called Masked Event Feature Modeling (MEFM) to learn the video-event representation for event detection.
Since the two sub-tasks share the same model architecture and parameters, we can alternately train the unified model with the three pre-training tasks, which is a simple, natural and effective method to strengthen the association between two sub-tasks. 
Experimental results on the ActivityNet Captions dataset~\cite{Krishna_2017_ICCV} show that even pre-trained on the same dataset without any extra data, our model significantly outperforms the state-of-the-art methods on both event detection and event captioning. In particular, benefiting from the new event detection framework, our model generates more diverse events similar to human annotations.
When pre-training the model on extra out-of-domain captioning data, our model achieves more additional gains, which demonstrates the effectiveness of our pre-training and fine-tuning framework.

The main contributions of this work are three-fold:
\parskip=0.1em
\begin{itemize}[itemsep=0.5pt,partopsep=0pt,parsep=\parskip,topsep=0.5pt]

    \item We transform the event detection sub-task into a sequence generation problem with temporal dependencies modeling between events, which makes our model directly generate an appropriate and low redundancy sequence of events for untrimmed videos.
    
    \item With the task format unification, we propose a unified video-event-text pre-training model with three pre-training tasks to effectively enhance the inter-task association between event detection and event captioning.
    
    \item Our model achieves new state-of-the-art dense video captioning results on the ActivityNet Captions dataset, and can be further promoted by pre-training on more out-of-domain video caption data.  
    
\end{itemize}

%% file: subsection/related_work.tex
\section{Related Works}
\noindent\textbf{Dense Video Captioning.}
The dense video captioning (DVC) task is first proposed by Krishna \etal \cite{Krishna_2017_ICCV} to localize and describe rich events in long untrimmed videos.
A two-stage framework is proposed, which first produces a large amount of event candidates via an event detection module, and then generates descriptions for each event.
Some following works separately improve the performance of two modules by producing less redundant event proposals~\cite{2020Streamlined} or introducing multi-modal features to enrich event representation~\cite{audio_dvc_single,Iashin_2020_CVPR_Workshops_single}.
However, handling event detection and captioning independently without any association has obvious limitations.
Therefore, some other following works~\cite{Wang_2018_CVPR_tda_cg,Deng_2021_CVPR,Zhou_2018_CVPR_MT,Li_2018_CVPR_dvc,Wang_2021_ICCV_pdvc} focus on enhancing the inter-task association between event detection module and event captioning module to improve the performance of DVC. 
Specifically, Li \etal \cite{Li_2018_CVPR_dvc} adopt  a new descriptiveness regression component to build a bridge between event detection and captioning modules, which measures the descriptive complexity of each event proposal, and adjusts the event proposal boundaries. 
Deng \etal \cite{Deng_2021_CVPR} reverse the predominant “detect-then-describe” fashion and propose a three-stage top-down framework, which first generates coarse-grained sentences, then aligns each sentence with event fragments, and finally refines the caption quality by a Dual-Path Cross Attention module. 
Wang \etal \cite{Wang_2021_ICCV_pdvc} extend the ``Object Query'' manner in DETR~\cite{DBLP,DETR} to the DVC task and decode the intermediate features and event query to produce event proposal and description simultaneously.

Although promising DVC results have been achieved in these methods, they fail to explore the temporal relationship between events for event detection, which leads to highly redundant event proposals.
Furthermore, the ``propose-then-select'' detection manner~\cite{DAPs_tep_anchor,Gao_2017_ICCV_tep_anchor,Shou_2016_CVPR_tep_anchor,Heilbron_2016_CVPR_tep_anchor} based on a large amount of event candidates is also computationally complex.
In this work, we convert the event detection task into a sequence generation problem, and unify the event detection and captioning into a unified framework, which makes the inter-task association more natural and effective.

\noindent\textbf{Pre-training for V+L Tasks.}
The pre-training and fine-tuning paradigm has demonstrated strong potentials in V+L~\cite{OSCAR_2020,uniter_2020_eccv,UniVL_2019_VL_P,Text_auto_2021,Su2020VL-BERT_VL_P,Huang_2021_CVPR_VL_P} cross-modal tasks recently.
Such as the Oscar~\cite{OSCAR_2020} model, first pre-trained on large-scale image-text pairs to learn a joint representation for vision and language, and then fine-tuned on several downstream tasks, achieves the state-of-the-art results on both vision-and-language understanding tasks (\emph{e.g.,} image-text retrieval~\cite{faghri2018vsepp}, visual question answering~\cite{Goyal_2017_CVPR_vqa_task}) and generation tasks (\emph{e.g.,} image captioning~\cite{vinyals2015showtell_img_cap}, novel object captioning~\cite{Agrawal_2019_ICCV_nocaps}). 
However, due to the coupling complexity of event detection and captioning, none of them have verified the effectiveness of multi-modal pre-training on the dense video captioning task.

%% file: subsection/method.tex
\section{Method}
Our unified model for dense video captioning following the pre-training and fine-tuning paradigm is illustrated in Figure~\ref{figure2}.
We unify the event detection and event captioning in one framework and treat the video, events and captions as three independent modalities.
Three pre-training tasks are proposed for the cross-modal representation learning, including Masked Language Modeling (MLM), Masked Video Feature Regression (MVFR), and Masked Event Feature Modeling (MEFM).
We first alternately pre-train the model with the three pre-training tasks and then fine-tune it for event detection and event captioning respectively.

\subsection{Model Architecture}

\noindent\textbf{Video Representation.}
Given an input video, we follow previous works~\cite{Wang_2021_ICCV_pdvc,Deng_2021_CVPR} to use C3D~\cite{Tran_2015_ICCV_c3d} and TSN~\cite{Tsn_wang_2019} to extract the raw video features in order to have a fair comparison in experimental evaluations. In addition, we also introduce semantic concept features (called CPT) to enrich the visual representation. 
Specifically, we employ a bidirectional LSTM~\cite{bi_lstm} as a multi-class multi-label video classifier to predict concept labels for each frame, which are the nouns and verbs extracted from corresponding annotated captions.
To enable the model to predict distinguishable fine-grained concept labels for different frames, we optimize the model with an additional event boundary prediction objective, which requires the model to predict whether the current frame is the start, middle or the end frame of an event. 
The hidden state from the last LSTM layer is used as the concept feature for each video frame.
Finally, we concatenate the concept feature with visual appearance features from C3D or TSN to represent the video as a sequence of frame-level features $V$ = $\{v_1, v_2, \cdots, v_N\}$, where $N$ is the number of frames in the video.
We employ a fully-connected layer to embed the video feature sequence $V$ into the same dimensionality as the word embedding, and add a learnable modality embedding $mode_v$ to represent the video modality. 

\begin{figure}[t]
\begin{center}
\centering
\includegraphics[width=\linewidth]{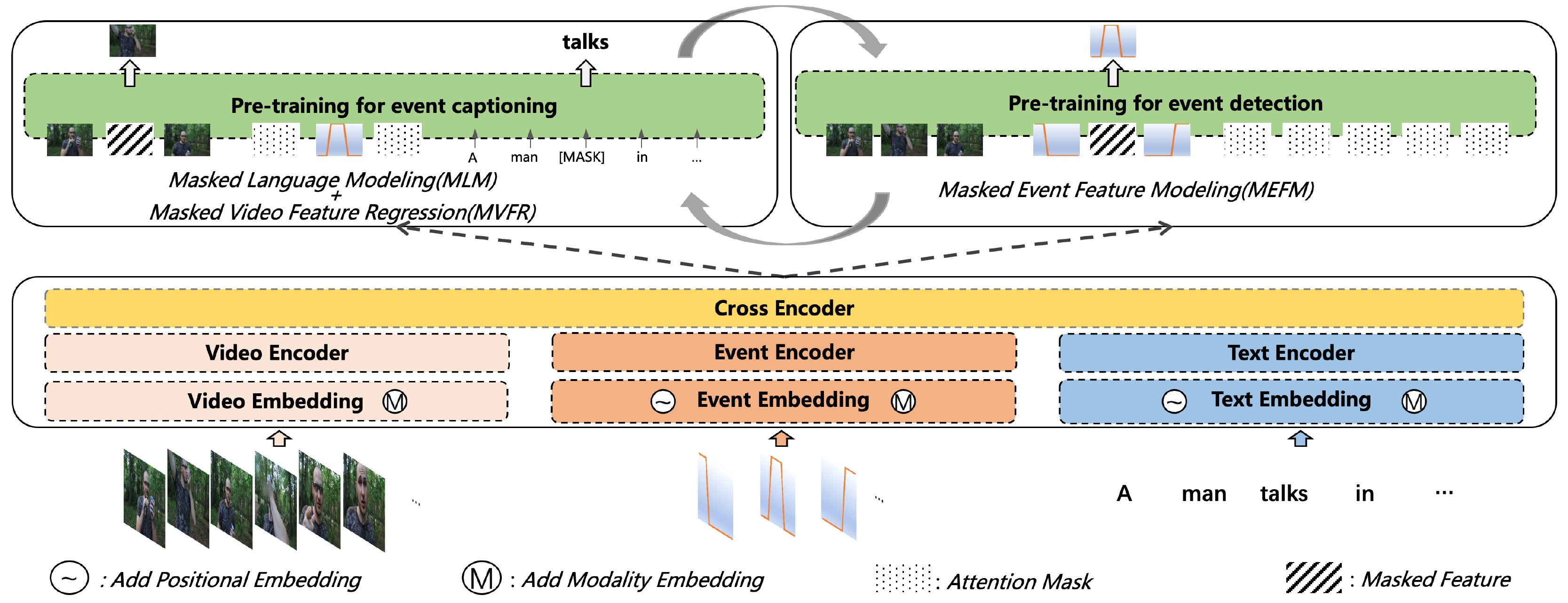}
\end{center}

\caption{Overview of our proposed model, which unifies three pre-training tasks for event detection and captioning. It contains three independent transformer encoders to capture the corresponding intra-modality context and a multi-layer cross encoder to encode the inter-modality context.}

\label{figure2}
\end{figure}

\noindent\textbf{Event Representation.}
For an event with timestamp $[start_i, end_i]$ in the video, we convert it into a $N$-dimensional binary feature vector, which is expressed as $e_i = \{x_1^i,x_2^i,\cdots,x_N^i\}$, where $N$ is the total number of video frames.
The value $x_t^i$ is set as 1 if the $t$-th frame is included in the event interval, otherwise it is set as 0.
In this way, we represent all the events in the video as a sequence of $E = \{e_1, e_2, \cdots, e_M\}$, where $M$ is the number of events in the current video. Similar to the video representation, we employ a fully-connected layer to embed the event sequence $E$ into the same dimensionality as the video and text embeddings, and add a learnable modality embedding $mode_e$ to represent the event modality. 
Furthermore, we add positional embedding to reserve the temporal order information of events in the video.

\noindent\textbf{Text Representation.}
For the caption text, we represent each caption with a sequence of word embeddings $T=\{t_1, t_2, \cdots, t_s\}$, where $s$ is the total number of words. 
We further add the positional embedding to keep the sequential information, and add a learnable modality embedding $mode_t$ to represent the text modality.

\noindent \textbf{Multimodal Transformer.}
Our model is based on a multi-stream architecture as illustrated in Figure~\ref{figure2}, where three independent transformer encoders are first applied on each modality for the intra-modality learning, and then a cross transformer encoder is employed to capture the inter-context information across different modalities.
We define the number of cross encoder layers as $L_c$ and the number of layers in the three single-modal encoders including the modality of video, event and text as $L_v$, $L_e$ and $L_t$ respectively.
The hidden size of all the transformer layers is denoted as $H$, and the number of self-attention heads is denoted as $A$.

\subsection {Proxy Tasks}

\noindent\textbf{Masked Language Modeling (MLM).}
To enable our model the ability for caption generation, we adopt MLM as one of the pre-training tasks like other Vision-and-Language (V+L) pre-training models~\cite{Text_auto_2021,uniter_2020_eccv,OSCAR_2020,UniVL_2019_VL_P,Su2020VL-BERT_VL_P,Huang_2021_CVPR_VL_P}.
The MLM task takes $V$, $E$ and $T$ as inputs, and predict the masked words in $T$ according to the corresponding video content of the current event timestamps $e_i$.
Since all the events in $E$ are input to the model while the caption $T$ is only corresponding to the current event, we restrict the attentions for other events $E_{\setminus i}$, so that the model can focus on the current event captioning. Specifically, we mask the cross-modal attention weights in the cross
encoder for other events $E_{\setminus i}$ to make the caption generation ignores other events.
Similar to the BERT~\cite{devlin2019bert}, we randomly mask out the words in $T$ with a probability of 15\%, and replace the masked ones $t_m$ with a special token [MASK] 80\% of the time, with another random word 10\% of the time and the original word 10\% of the time.
The goal of this task is to predict the masked words based on the context information from the whole video content, the current event boundary $e_i$, and the surrounding captioning words by minimizing the negative log-likelihood as follows:
\begin{equation}
    \mathcal{L}_{\rm MLM} = - \mathbb{E}_{(V,e_i,T)\sim\mathcal{D}} \log p(t_m|t_{\setminus m},e_i,V;\Theta),
\end{equation}
where $\Theta$ denotes all trainable parameters, $\mathcal{D}$ denotes the whole training set, and $t_m$ denotes the masked words in $T$.

\noindent\textbf{Masked Video Feature Regression (MVFR).}
In contrast to the MLM task which predicts the masked caption words according to the video content, we also introduce the MVFR task to reconstruct video features based on the description.
The input for MVFR task is exactly the same as the MLM task.
Suppose $T$ is the description for the event $e_i$, we randomly choose some video frames in $e_i$ for prediction.
Specifically, we randomly mask 15\% of the features from $\{V_{i}\}_{i=s}^{e}$, where $v_s$ and $v_e$ are the start and end video frame in $e_i$ respectively.
Each masked feature $v_m$ is replaced by a special feature vector [MASK], which is an all-zeros vector with the same dimensionality as the original video feature $r(v_m)$.
The hidden state of $v_m$ from the cross transformer encoder is input to a FC layer to predict the original video feature denoted as $p({v_m})$, according to the remaining video features $V_{\setminus m}$ and the event description $T$.
We adopt the L2 regression loss to reduce the distance between $p({v_m})$ and $r({v_m})$ as follows:
\begin{align}
    & \mathcal{L}_{\rm MVFR} = \mathbb{E}_{(V,e_i,T)\sim\mathcal{D}} FC(v_m|v_{\setminus m},e_i,T;\Theta), \\
    & FC(v_m|v_{\setminus m},e_i,T;\Theta) =  \sum_{j=1}^N \| p(\mathbf{v}_\mathbf{m}^{(j)}) - r(\mathbf{v}_\mathbf{m}^{(j)}) \|_{2}^2,
\end{align}
where $N$ is the dimension of video features.
We share the FC prediction layer with the video feature embedding layer.

\noindent\textbf{Masked Event Feature Modeling (MEFM).}
To predict event boundaries in a sequence generation manner, we propose a new pre-training task called Masked Event Feature Modeling (MEFM).
Unlike above two pre-training tasks, the MEFM task takes $V$, $E$ as inputs. 
We randomly mask event embeddings in $E$ with a probability of 15\%, and replace the masked one $e_m$ with a special feature vector [MASK], which is an all-zeros vector.
The model is required to predict the event boundaries of $e_m$ according to the surrounding events in $E$ and the video content.
We apply a FC layer on the output of cross transformer encoder as a $N$-dimensional binary classifier, where $N$ is the length of video frame sequences.
The training loss can be defined as follows:
\begin{align}
    & \mathcal{L}_{\rm MEFM} = \mathbb{E}_{(V,E)\sim\mathcal{D}}f(e_m|e_{\setminus m},V;\Theta), \\
    & f(e_m|e_{\setminus m},V;\Theta) = \sum_{i=1}^N [l_{i} \log x_i + (1-l_{i}) \log (1-x_i)],
\end{align}
where $l_{i}$ $\in$ [0, 1] indicates whether the \textit{i}-th video frame is included in the masked event $e_m$ interval, and $x_i$ denotes the predicted probability.

\subsection {Pre-Training and Fine-Turning for DVC Task} 
As shown in Figure~\ref{figure2}, we pre-train the model for event detection and event captioning in a unified framework. 
However, the pre-training tasks of MLM and MVFR for event captioning take three modalities (video, event, text) as input, while the MEFM task for event detection takes two modalities without the text modality as input, making it problematic to optimize the three objectives jointly.
Therefore, we train the two sub-tasks alternatively.
Specifically, we divide the batch data into two categories: the batch with three modalities input (for MLM, MVFR) named as $B_\textit{three}$, and the batch with two modalities input (for MEFM) named as $B_\textit{two}$. 
We introduce a hyper-parameter $\lambda$ to represent the probability of choosing batch $B_\textit{three}$, and thus ($1$-$\lambda$) denotes the probability of choosing batch $B_\textit{two}$. 
When the batch $B_\textit{three}$ is fed to the model, the training loss is defined as $\mathcal{L}=\mathcal{L}_{\rm MLM}+\mathcal{L}_{\rm MVFR}$, while when batch $B_\textit{two}$ is fed to the model, the loss is defined as $\mathcal{L}=\mathcal{L}_{\rm MEFM}$.

After pre-training, we fine-tune the model for the event detection sub-task (called ED) and event captioning sub-task (called EC) respectively. 
The two downstream tasks are similar since they both follow the auto-regressive manner~\cite{cornia2020meshed_captioning,huang2019attention_captioning} for generation. Therefore, we fine-tune the model in a similar way for the two sub-tasks.
Specifically, we adapt our bi-directional pre-trained model to a uni-directional generator by constraining the self-attention mask of the text/event sequence to avoid seeing future items. 
Similar to the MLM and MEFM pre-training tasks, we randomly choose 15\% of word/event features and replace them with the [MASK] token/special all-zeros vector for prediction.
Note that the event detection task is to generate event sequences with the whole video as input, while the event captioning task generates captions for the corresponding event according to the whole video and event embedding $e_i$.
Therefore, the fine-tuning objective for ED and EC tasks can be expressed as follows:
\begin{align}
    & \mathcal{L}_{\rm ED} = - \mathbb{E}_{(V,E)\sim\mathcal{D}} \log p(e_m|e_{<m},V;\Theta), \\
    & \mathcal{L}_{\rm EC} = - \mathbb{E}_{(V,e_i,T)\sim\mathcal{D}} \log p(t_m|t_{<m},e_i,V;\Theta).
\end{align}

In the inference phase, we follow the ``detect-then-describe'' pipeline.
At the stage of event prediction, we first input the whole video frame sequence and a special ``start event'' vector to the model. 
Then, we start to generate the event sequence one by one via feeding a [MASK] vector and sampling the predicted event feature from the $N$-dimensional binary classification layer. 
The predicted event feature vector is then used to replace the previous [MASK] vector, and a new [MASK] vector is fed to the model for the next event generation until the special ``end event'' vector is predicted. 
Finally, following the rule that the first 1-value in the event vector is regarded as the start time of an event in the video, and the last 1-value in the event vector is regarded as the end time of the event in the video, we translate the event vector into the event timestamp format.
After predicting the event sequence for the untrimmed video, we generate corresponding captions for each predicted event. 
We first input the whole video, the current event embedding and the start [SOS] token to the model. Then, we follow the same auto-regressive sequence generation process to generate the captions until the end [EOS] token is predicted.

%% file: subsection/experiment.tex
\section{Experiments}

\subsection{Experimental Settings}
\noindent \textbf{Dataset.}
We conduct experiments on the benchmark ActivityNet Captions dataset~\cite{Krishna_2017_ICCV}, which contains 19,994 videos with an average of 3.65 event proposals per video. We follow the official split with 10009/4925/5044 videos for training, validation, and test. 
Furthermore, to demonstrate the ability of our model to benefit from more out-of-domain captioning data, we further evaluate our model pre-trained with other non-dense video captioning datasets, including MSRVTT~\cite{Xu_2016_CVPR_MSRVTT}, TGIF~\cite{Li_2016_CVPR_TGIF} and VATEX~\cite{Wang_2019_ICCV_VATEX} datasets.

\noindent \textbf{Implementation details.}
For fair comparisons with the state-of-the-art methods, we use the same video features as PDVC~\cite{Wang_2021_ICCV_pdvc}, including the C3D~\cite{Tran_2015_ICCV_c3d} features provided by PDVC~\cite{Wang_2021_ICCV_pdvc} and TSN~\cite{Tsn_wang_2019} features provided by MT~\cite{Zhou_2018_CVPR_MT}.
The max length of video frames $V$ is set as 100.
We set the layer number of independent encoders $L_v = L_e = L_t = 1$, the layer number of cross encoder $L_c = 4$, the hidden size $H = 512$, and the head number $A = 8$. 
When pre-training the model for event detection, the hyper-parameter $\lambda$ for choosing different batches is set as 1/3 on both C3D and TSN features. When pre-training the model for event captioning, it is set as 1/2 on C3D features and 3/4 on TSN features. We use Adam~\cite{Adam} as the optimizer and train all the model parameters from scratch.

\noindent \textbf{Metrics.}
We evaluate our method from three aspects. \textbf{(1)} To evaluate the performance of event detection, we use the evaluation tool provided by the 2018 ActivityNet Captions Challenge\footnote{https://github.com/ranjaykrishna/densevid\_eval} (called Evaluator2018), which computes the average precision and average recall against the ground-truth events across temporal IoU (tIoU) at [0.3, 0.5, 0.7, 0.9]. Moreover, we also compute the self-tIoU between the detected events to evaluate the event diversity. \textbf{(2)} To purely evaluate the event captioning ability, we report the captioning results based on the ground-truth events with classic captioning metrics, including BLEU~\cite{papineni:bleu}, METEOR~\cite{lavie2005meteor} and CIDEr~\cite{Vedantam_2015_CVPR_cider}.
We also use the Evaluator2018 to compute these metrics and report the results with tIoU threshold of 0.9.
\textbf{(3)} To evaluate the performance of dense video captioning, the captioning performance based on the \textbf{\textit{generated}} events, we use SODA\footnote{https://github.com/fujiso/SODA}~\cite{SODA_2020} as the evaluation metric, which is a new evaluation metric proposed for dense video captioning to overcome some of the limitations of previous metrics.

\begin{table}[t]
\caption{The DVC performance measured by different evaluation tools on the original results and results with four adverse operations. SODA$_{old}$ is the old version of SODA implementation and SODA$_{mr}$ is the new implementation with multiple references.}
\label{tab:eval_tool}
\centering
\small
\begin{tabular}{l | c | c | c c c|c|c}
\toprule
\multirow{2}{*}[-0.5ex]{Operation} & \multirow{2}{*}[-0.5ex]{\shortstack{Avg\\Recall}}& \multirow{2}{*}[-0.5ex]{\shortstack{Avg\\precision}}& \multicolumn{3}{c|}{Evaluator2018} &  \multirow{2}{*}[-0.5ex]{SODA$_{old}$} & \multirow{2}{*}[-0.5ex] {SODA$_{mr}$} \\ 
\cmidrule{4-6}
& & & BLEU@4 & CIDEr& METEOR & \\
\midrule
\emph{Original} & 59.00 & 60.32 & 1.45 & 26.92 & 7.33 & 5.29 &7.28 \\ \midrule
\emph{Increase} & 59.00 & 60.08 & 1.57 & 25.94 & 7.48 & 5.04 &6.91 \\
\emph{Reduce} & 53.75 & 60.21 & 1.55 & 27.99 & 7.44 & 5.06 &6.99 \\
\emph{Exchange} & 53.75 & 59.99 & 1.66 & 26.45 & 7.58 & 4.81 &6.61 \\
\emph{Extreme} & 20.63 & 58.76 & 2.45 & 14.94 & 8.52 & 2.98 &4.64 \\
\bottomrule
\end{tabular}

\end{table}

To verify that the general video captioning metrics are not appropriate enough for the dense video captioning evaluation, we carefully design an experiment to compare the classic evaluation metrics provided in Evaluator2018 with the newly proposed SODA metric.
We first compute the scores of captions for events in different intervals of the video, and observe that the first event of the video usually gets higher captioning scores than other locations.
Based on such observation, we propose four simple operations, including \emph{Increase}, \emph{Reduce}, \emph{Exchange} and \emph{Extreme}, to modify our dense video captioning results, and then show the variations of scores computed by different evaluation metrics. 
Given a submission file with the best results of our model, the \emph{Increase} operation randomly copies the first event and its corresponding caption in the submission file with 40\% probability for each video.
The \emph{Reduce} operation randomly removes the $i$-th (where $i>1$) event and its corresponding caption in the submission file with 15\% probability.
The \emph{Exchange} operation is the combination of \emph{Increase} and \emph{Reduce} operations, and the \emph{Extreme} operation removes the $i$-th (where $i>1$) event and its corresponding caption with 100\% probability.
We run the four operations with different random seeds for three times and report the average results in Table~\ref{tab:eval_tool}.
Although we intuitively expect that the above four operations should adversely affect the dense video captioning results, the scores evaluated by the Evaluator2018 surprisingly increase significantly from 7.33 to 8.52 on the METEOR metric. On the contrary, SODA correctly reflects the captioning quality with the score constantly decreased. 
It is because the Evaluator2018 fails to penalize redundant and non-recalled events. Therefore, simply repeating the first event caption with 40\% probability can significantly improve the BLEU@4 and METEOR scores by 8\% and 2\% respectively, although the results actually do not have any substantial improvements.
Therefore, we consider SODA as the main evaluation metric for dense video captioning. 
In this work, we use two versions of SODA to evaluate our DVC performance.
The SODA$_{old}$ is commonly used in previous methods, which computes the score based on two references independently and reports the averaged score.
The SODA$_{mr}$ however computes the score based on multiple references simultaneously, which are more accurate.

\subsection{Comparison with State-of-the-art Methods}

\begin{table*}[t]
\caption{Event detection results (using C3D feature) on the ActivityNet Captions validation set. sIoU refers to self-tIoU. The best results are in bold and the second best are underlined.}
\label{tab:ed_task}
\centering
\small
\begin{tabular}{l | c c c c c| c c c c c|c}
\toprule
\multirow{2}{*}[-0.5ex]{Method} & \multicolumn{5}{c|}{Recall} & \multicolumn{5}{c|}{Precision} & \multirow{2}{*}[-0.5ex]{ sIoU} \\ 
\cmidrule{2-11}
&0.3&0.5&0.7&0.9 &avg &0.3&0.5&0.7&0.9&avg& \\
\midrule
MFT\cite{Xiong_2018_ECCV_mft} & 46.18 & 29.76 & 15.54 & 5.77 & 24.31 & 86.34 & 68.79 & 38.30 & 12.19 & 51.41 & - \\
SDVC\cite{2020Streamlined} & \underline{93.41} & \underline{76.40} & 42.40 & 10.10 & \underline{55.58} & 96.71 & 77.73 & \underline{44.84} & 10.99 & 57.57 & - \\
PDVC\cite{Wang_2021_ICCV_pdvc} &  89.47 & 71.91 & \underline{44.63} & \textbf{15.67} & 55.42 & \textbf{97.16} & \underline{78.09} & 42.68 & \textbf{14.40} & \underline{58.07} & 0.19 \\
\textbf{Ours} &  \textbf{94.68} & \textbf{80.95} & \textbf{47.84} & \underline{12.54} & \textbf{59.00} & \underline{96.97} & \textbf{80.80} & \textbf{50.15} & \underline{13.37} & \textbf{60.32} & 0.02 \\
\bottomrule
\end{tabular}
\end{table*}

We compare our model with both types of baseline methods, including the ones that tackle event detection and captioning separately, such as MFT~\cite{Xiong_2018_ECCV_mft}, SDVC~\cite{2020Streamlined} and ECHR~\cite{wang2021echr}, and the state-of-the-art models that exploit interactions between the two sub-tasks, such as DCE~\cite{Krishna_2017_ICCV}, TDA-CG~\cite{Wang_2018_CVPR_tda_cg}, DVC~\cite{Li_2018_CVPR_dvc}, MT~\cite{Zhou_2018_CVPR_MT}, PDVC~\cite{Wang_2021_ICCV_pdvc} and SGR~\cite{Deng_2021_CVPR}.

\noindent \textbf{Event detection results.} 
As shown in Table \ref{tab:ed_task}, our proposed event sequence generation model outperforms previous event detection methods by a large margin.
Although pre-trained on the same dataset without any extra data, our model achieves significant improvements over previous best results at most of tIoU thresholds, with the average recall score improved from 55.58\% to 59.00\% and the average precision score improved from 58.07\% to 60.32\% .
Furthermore, our model generates 2.94 events per video on average with the self-tIoU of 0.02, which shows a greater improvement on reducing the event redundancy than traditional event detection methods. It is also much closer to the ground-truth events which have an average self-tIoU of 0.05.
In addition to achieving much better results on the event detection accuracy and diversity, our model is also more efficient than previous methods.
The PDVC~\cite{Wang_2021_ICCV_pdvc} model needs to train an additional predictor for event prediction and 
SDVC~\cite{2020Streamlined} first adopts an extra model (SST~\cite{buch_sst}) to extract 1000 event proposals and obtains M candidate proposals with Non-Maximum Suppression. Then, SDVC selects the final proposals from the M candidates in an auto-regressive manner.
On the contrary, our model directly generates an appropriate number of events from the raw input video features, which is one-stage without error accumulation and with much less computational cost. 

\begin{table*}[t]
\caption{Event captioning results on the ActivityNet Captions validation set. SODA$_{old}$ is the old version of SODA implementation and SODA$_{mr}$ is the new implementation with multiple references.}
\label{tab:vc_task}
\centering
\small
\begin{tabular}{l | c | c c c | c c}
\toprule
\multirow{2}{*}[-0.5ex]{Method} & \multirow{2}{*}[-0.5ex]{Features} &\multicolumn{3}{c|}{{with Ground-Truth proposals}} & \multicolumn{2}{c}{{with generated proposals}} \\ 
\cmidrule{3-7}
&  & BLEU@4& METEOR & CIDEr & SODA$_{old}$ & SODA$_{mr}$  \\
\midrule
DCE\cite{Krishna_2017_ICCV} & C3D & 1.60 & 8.88 & 25.12 & - & - \\
TDA-CG\cite{Wang_2018_CVPR_tda_cg} & C3D & - & {9.69} & - & -& -  \\
DVC\cite{Li_2018_CVPR_dvc} & C3D& 1.62 & 10.33 & 25.24 & -& -  \\
ECHR\cite{wang2021echr} & C3D& 1.96 & 10.58 & 39.73 & 3.22 & -  \\
PDVC\cite{Wang_2021_ICCV_pdvc} & C3D& 2.64 & 10.54 & 47.26 & 5.26 & 7.14 \\
\textbf{Ours} & C3D & \textbf{2.67} & \textbf{11.01} & \textbf{52.42} & \textbf{5.29} &  \textbf{7.28}\\
\midrule
MT\cite{Zhou_2018_CVPR_MT} & TSN & 2.71 & 11.16 & 47.71 & 4.02 & - \\
SGR\cite{Deng_2021_CVPR} & TSN & - & - & - & 5.29 & -\\
PDVC\cite{Wang_2021_ICCV_pdvc} & TSN & \textbf{3.07} & 11.27 & 52.53 & 5.44 & 7.42\\
\textbf{Ours} & TSN & 2.90 & \textbf{11.43} & \textbf{54.75} & \textbf{5.49} & \textbf{7.61}\\
\bottomrule
\end{tabular}

\end{table*}

\noindent \textbf{Event captioning results.} 
Table~\ref{tab:vc_task} reports the dense video captioning results of different models on the ActivityNet Captions validation set. 
To have a fair comparison with the state-of-the-art methods, we train our model with C3D and TSN visual features respectively.
When inferring based on the ground-truth event proposals, our model trained by C3D features improves the BLUE@4 from 2.64 to 2.67, the METEOR from 10.58 to 11.01, and the CIDEr from 47.26 to 52.42.
Similar improvement trends are obtained with the TSN features on METEOR and CIDEr as well.
It demonstrates the effectiveness of our model for event captioning.
When inferring based on the generated proposals, we follow the ``detect-then-describe'' pipeline, which first predicts the event sequence and then describes each event clip.
We evaluate our model with the more appropriate evaluation metric SODA for dense video captioning (SODA~\cite{SODA_2020}), and report two scores, including SODA$_{old}$ and SODA$_{mr}$.
As shown in Table~\ref{tab:vc_task}, our model with C3D features achieves the best SODA$_{old}$ score and SODA$_{mr}$ score, outperforming all previous methods. 
The performance of our model with more advanced TSN features surpasses all previous methods as well.

\subsection{Ablation Studies}

\begin{table}[t]
 \caption{Ablation study to demonstrate the effectiveness of different components for event captioning.}
        \label{table:abl_vc}
        \centering
        \begin{tabular}{c | c c c | c | c c c}
                \toprule
                 \multirow{2}{*}[-0.5ex]{Row} & \multicolumn{4}{c|}{Components} & \multirow{2}{*}[-0.5ex]{BLEU@4} & \multirow{2}{*}[-0.5ex]{METEOR} & \multirow{2}{*}[-0.5ex]{CIDEr} \\
                 \cmidrule{2-5}
                & MLM & MVFR & MEFM & CPT & & &  \\
                \midrule
                 1 &  &  &  &  & 2.22 & 9.98 & 47.7 \\
                 2 & \checkmark &  &  &  & 2.17 & 10.20 & 49.05 \\
                 3 & \checkmark & \checkmark & &  & 2.38 & 10.48 & 51.08 \\
                 4 & \checkmark &\checkmark  & \checkmark &  & 2.59 & 10.94 & 52.13 \\
                 5 & \checkmark &\checkmark  & \checkmark  & \checkmark  & 2.67 & 11.01 & 52.42 \\
                \bottomrule
            \end{tabular}
\end{table}

\begin{table}[t]
 \centering
        \caption{Ablation study to demonstrate the effectiveness of different components for event detection.}
        \label{table:abl_ed}
        \begin{tabular}{c | c c c |c | c c }
                \toprule
                \multirow{2}{*}[-0.5ex]{Row} & \multicolumn{4}{c|}{Components} & \multirow{2}{*}[-0.5ex]{Avg@Recall} &
                \multirow{2}{*}[-0.5ex]{Avg@Precision}\\
                \cmidrule{2-5}
                  & MEFM & MLM & MVFR & CPT &  & \\
                \midrule
                 1 &  &  &  &  & 54.94 & 57.95 \\
                 2 & \checkmark &  &  &  & 56.49 & 58.12 \\
                 3 & \checkmark & \checkmark & & & 58.00 & 59.93 \\
                 4 & \checkmark & \checkmark & \checkmark &  & 58.32 & 59.95 \\
                 5 & \checkmark &\checkmark & \checkmark   & \checkmark  & 59.00 & 60.32 \\
                \bottomrule
            \end{tabular}

\end{table}

\begin{figure*}[ht]
	\begin{center}
		\includegraphics[width=0.93\linewidth]{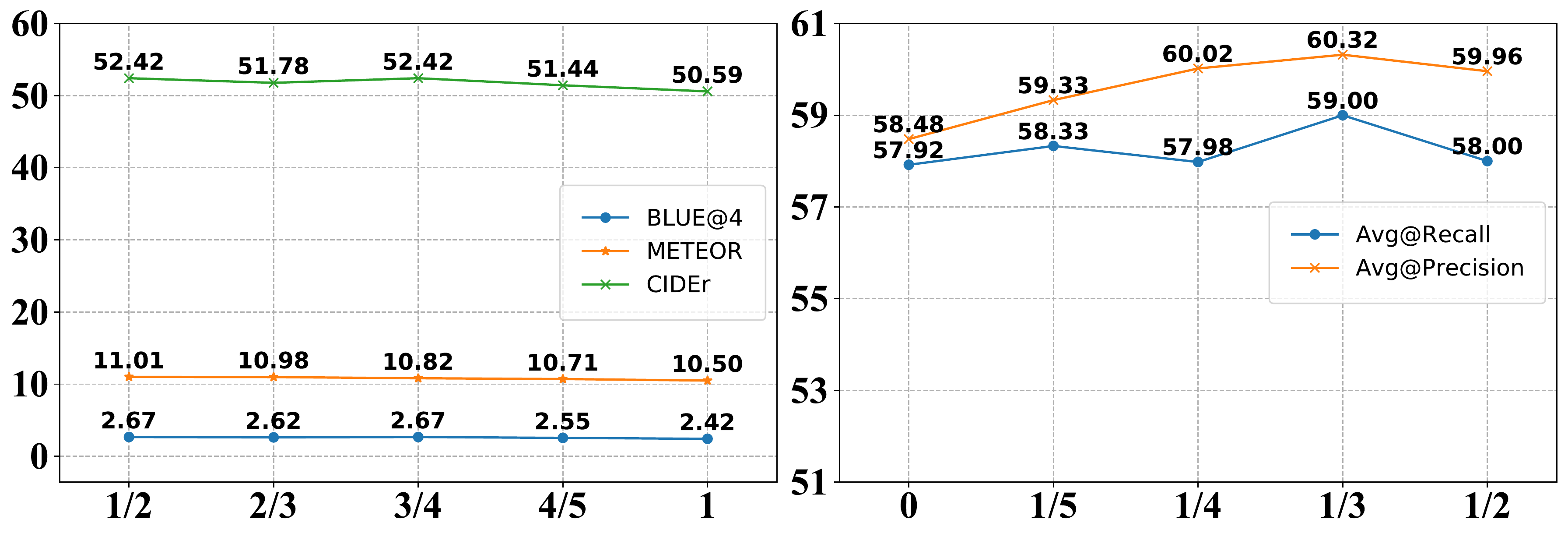}
	\end{center}

	\caption{Performance variations with different trade-off parameter $\lambda$. The left and right sub-figures show the event captioning and event detection results respectively. }

	\label{fig:abl_vary}
\end{figure*}

In this section, we ablate our model to demonstrate the effectiveness of different components for event captioning and event detection respectively.

Table~\ref{table:abl_vc} shows the ablation results of our model for the event captioning. To purely analyse the caption generation qualities, we use the  ground-truth video events to avoid the impact of event detection quality.
As shown in Table~\ref{table:abl_vc}, the model pre-trained only with MLM task (Row 2) has outperformed the non-pretrain baseline (Row 1) even without any extra data for pre-training.
When combining with other pre-training tasks, including MVFR and MVFR+MEFM (Row 3 and 4), our model achieves significant improvements on the captioning metrics, which demonstrates the effectiveness of the proposed pre-training tasks.
Specifically, the improvement brought by MEFM task is the most significant, which improves the BLEU@4 from 2.38 to 2.59, the METEOR from 10.48 to 10.94 and the CIDEr from 51.08 to 52.13.
It shows that the event captioning task can benefit from the event detection task, and demonstrates the advantages of our proposed unified framework for the inter-task interaction.

The same trend can also be found in the Table~\ref{table:abl_ed}.
The pre-trained model significantly outperforms the non-pretrain baseline in Row 1.
Adding the MLM and MVFR tasks in the pre-training stage greatly improves the event detection results, with the average recall improved from 56.49 to 58.32 and the average precision improved from 58.12 to 59.95.
It shows that the event detection task can also benefit from the event captioning task.
Besides different pre-training tasks, enhancing the video representation with semantic concept features also brings additional gains for both the event captioning and event detection tasks, as shown in Row 5 vs. Row 4 of the Table~\ref{table:abl_vc} and Table~\ref{table:abl_ed}.
The surprising performance improvement by the semantic concept features (short for CPT) in event detection (improved from 58.32 to 59.00 on average recall and 59.95 to 60.32 on average precision) further demonstrates that comprehensive semantic understanding of video frames can greatly help event boundary detection.
This is also the reason why the event detection and event captioning can help each other, because they are both based on the semantic understanding of videos.
Finally, combining all the components together in our model achieves the best performance for both event detection and event captioning (Row 5).

We also ablate the influence of the hyper-parameter $\lambda$ in our model in Figure~\ref{fig:abl_vary}. We observe that enhancing inter-task association in different degrees  all outperform  the model without inter-task interaction (where $\lambda = 0 \ or \ 1$) for  both  the  event  captioning  and  event  detection.
With gradually increasing $\lambda$ from 1/2 to 1, the METEOR score continues to decline, while the CIDEr and BLUE@4 scores fluctuate slightly. We observe that modulating hyper-parameter $\lambda$ to 1/2 obtains the best performance for event captioning. While when pre-training for event detection, the best performance is achieved when hyper-parameter $\lambda$ is set as 1/3.

\subsection{Pre-training with Out-of-domain Data}

\begin{figure*}[t]
	\begin{center}
		\includegraphics[width=0.93\linewidth]{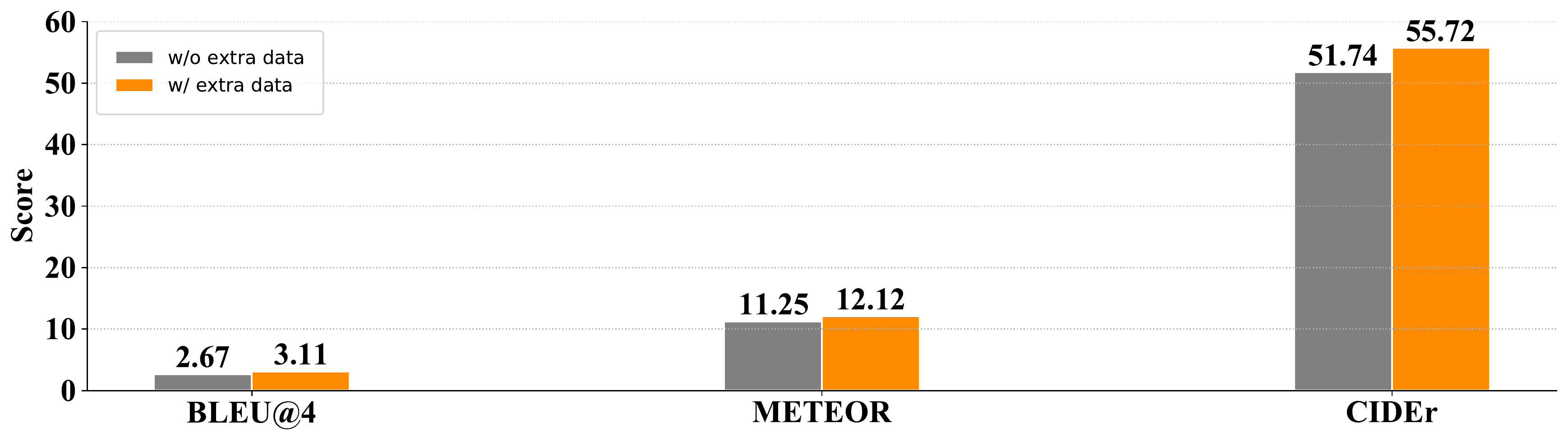}
	\end{center}
	\caption{Event captioning w/ and w/o extra pre-training data.}
	\label{fig:vc_extra_task}
\end{figure*}

Due to the advantage of pre-training and fine-tuning framework, our model can benefit from more out-of-domain data besides the ActivityNet dataset.
Specifically, we pre-train the model with conventional non-dense video captioning datasets including MSRVTT~\cite{Xu_2016_CVPR_MSRVTT}, TGIF~\cite{Li_2016_CVPR_TGIF} and VATEX~\cite{Wang_2019_ICCV_VATEX}, which results in 676K additional video-caption pairs in total.
Since the conventional video captioning datasets do not contain event annotations, we pre-train the model only with the MLM and MVFR tasks on these out-of-domain data, and fine-tune it on the ActivityNet Captions training set to adapt to the dense video captioning task.
Figure~\ref{fig:vc_extra_task} shows the results of our model pre-trained with extra data.
Compared with the model pre-trained only on the ActivityNet dataset, the additional out-of-domain data brings significant additional gains on all the captioning metrics, e.g., improving BLEU@4, METEOR and CIDEr by more than 0.4, 0.8 and 4 points respectively.
It demonstrates the merit of our unified pre-training based model framework, which enables the dense video captioning task to benefit from conventional non-dense video captioning datasets.

%% file: subsection/conclusion.tex
\section{Conclusion}
In this work, we define the event detection task as a sequence generation problem to fully exploit the temporal relationship between events for more accurate and diverse event detection.
Benefiting from the unification of event detection and event captioning sub-tasks, we propose a unified dense video captioning model based on pre-training and fine-tuning framework.
We design a new ``event'' modality and propose three pre-training tasks to interact the event detection and event captioning sub-tasks naturally.
Experimental results on the ActivityNet Captions dataset show that our model significantly outperforms the state-of-the-art methods on both event detection and event captioning, and achieves additional gains when leveraging more out-of-domain data.